%% file: main.tex
\documentclass[runningheads]{llncs}
\usepackage{graphicx}

\input{newcommand}

\usepackage{booktabs} %
\usepackage{amsmath,amssymb,amsfonts}
\usepackage{mathtools}
\usepackage{pdfpages}

\usepackage{tikz}
\usepackage{comment}
\usepackage{color}

\usepackage[accsupp]{axessibility}  %
\usepackage{bbding}

\makeatletter
\def\thanks#1{\protected@xdef\@thanks{\@thanks
        \protect\footnotetext{#1}}}
\makeatother

\begin{document}
\pagestyle{headings}
\mainmatter
\def\ECCVSubNumber{3726}  %

\title{UnrealNAS: Can We Search Neural Architectures with Unreal Data?} %

\titlerunning{UnrealNAS: Can We Search Neural Architectures with Unreal Data?}
\author{Zhen Dong\inst{1}$^*$ \and
Kaicheng Zhou\inst{1}$^*$ \and
Guohao Li\inst{4}$^*$ \thanks{Equal contribution. Corresponding author: Shanghang Zhang, E-mail: {\tt\small shanghang@pku.edu.cn}} \and
Qiang Zhou\inst{3} \and
Mingfei Guo\inst{2} \and
Bernard Ghanem\inst{4} \and
Kurt Keutzer\inst{1} \and
Shanghang Zhang\inst{2}~\Envelope}
\authorrunning{Z. Dong et al.}
\institute{University of California, Berkeley 
\and
Peking University
\and
Tsinghua University
\and
King Abdullah University of Science and Technology\\
\email{\{zhendong, caseyzhou, keutzer\}@berkeley.edu, \{guohao.li,Bernard.Ghanem\}@kaust.edu.sa, bamboosdu@gmail.com, \{mingfeiguo,shanghang\}@pku.edu.cn}}
\maketitle

\input{Section/abstract}

\input{Section/introduction}

\input{Section/related}

\input{Section/method_analysis}

\input{Section/experiment}

\input{Section/ablation}

\input{Section/conlusion}

\clearpage
\bibliographystyle{splncs04}
\bibliography{egbib}

\onecolumn

\clearpage
\input{Section/supp}

\end{document}

%% file: newcommand.tex
\usepackage{braket}
\usepackage{xcolor}
\usepackage{colortbl}
\usepackage{xspace}

\newcommand{\ie}{\emph{i.e.}\xspace}
\newcommand{\vs}{\emph{vs.}\xspace}

\newcommand{\figLabel}{Figure\xspace}
\newcommand{\eqnLabel}{Equation\xspace}

\newcommand{\mysection}[1]{\vspace{0pt}\noindent\textbf{#1.}}

\newcommand{\savespace}{\vspace{0pt}}

\newcommand\gc{ \rowcolor{gray!40}}

%% file: Section/abstract.tex
\begin{abstract}
Neural architecture search (NAS) has shown great success in the automatic design of deep neural networks (DNNs). However, the best way to use data to search network architectures is still unclear and under exploration. 
Previous work~(\cite{liu2020labels,zhang2021neural}) has analyzed the necessity of having ground-truth labels in NAS and inspired broad interest. In this work, we take a further step to question whether real data is necessary for NAS to be effective. The answer to this question is important for applications with limited amount of accessible data, and can help people improve NAS by leveraging the extra flexibility of data generation. 
To explore if NAS needs real data, we construct three types of unreal datasets using: 1) randomly labeled real images; 2) generated images and labels; and 3) generated Gaussian noise with random labels. 
These datasets facilitate to analyze the generalization and expressivity of the searched architectures.
We study the performance of architectures searched on these constructed datasets using popular differentiable NAS methods. Extensive experiments on CIFAR, ImageNet and CheXpert~\cite{irvin2019chexpert} show that the searched architectures can achieve promising results compared with those derived from the conventional NAS pipeline with real labeled data, suggesting the feasibility of performing NAS with unreal data.
\end{abstract}

%% file: Section/introduction.tex
\section{Introduction}
Advances in designing neural architectures have substantially contributed to the success of deep learning. 
In addition to manual design based on domain knowledge, neural architecture search (NAS) algorithms enable the automated search for effective network architectures. 
Approaches to exploring the model-architecture design space
include reinforcement learning \cite{zoph2016neural,zoph2018learning,pham2018efficient,tan2019mnasnet}, 
evolutionary algorithms \cite{liu2017hierarchical,real2019regularized,guo2020single}, sampling-based methods~\cite{ru2020neural,zela2018towards,dong2021hao}, and differentiable methods \cite{liu2018darts,xie2018snas,dong2019search,chen2019progressive,xu2019pc,chu2019fairdarts,li2019sgas,Zela2020Understanding}.
Although current approaches have demonstrated their effectiveness,
the best way to use data to  
evaluate design points and search effectively for neural-architectures in a computationally efficient manner
is still under exploration. 
Some NAS algorithms partially train the networks in the design
space under
consideration on the training set (or a subset of training set) and use the performance on the validation set as a criterion. 
This is valid with the assumption that the best-performing networks during the partial training process will consistently outperform the others during the whole training process. Other NAS algorithms take advantage of transferability and evaluate the network on proxy tasks such as  a similar but smaller dataset (a common example is CIFAR-10), or a simpler task comparing to the original one. The success of these methods shows a strong correlation of the network performance among a variety of tasks, suggesting there are some data-agnostic factors of an architecture that determine its effectiveness on learning. In this paper, we aim to analyze such factors and investigate what tasks and datasets can be used to evaluate them.

It has been shown in previous literature that the performance of a neural network architecture depends on its effective capacity rather than solely on its expressivity~\cite{liu2020labels,zhang2021neural}. In addition to aforementioned proxy methods, UnNAS~\cite{liu2020labels} proposed to use pretext tasks in self-supervised learning as an evaluation of the effective capacity. Specifically, UnNAS replace the original supervised learning objective in DARTS (Differentiable Architecture Searching)~\cite{liu2018darts} with rotation prediction, colorization and jigsaw puzzles, which leads to decent results with only self-supervised labeling. RLNAS~\cite{zhang2021neural} made a step further by randomly labelling the dataset and selecting neural architectures based on the ease-of-convergence hypothesis, which assumes a network with higher convergence speed tends to have a higher effective capacity. Intrigued by these works exploring characteristics of tasks and training, in our work, we propose UnrealNAS to investigate the network generalization and expressivity of the searched architectures
from the perspective of input data. 
Specifically, we want to explore if the original real data is a necessity to search good neural architectures? How will different forms of input data affect the searching results? What relationship between input data and labelling works the best for evaluating the effective capacity? 
In order to answer these questions, we come up with three experimental settings with unreal data and labels: 1) real data with random labels; 2) generated data with random labels; and 3) random noise with random labels. 
Based on the assumption that difficult tasks are beneficial to evaluate the effective capacity, we analyze the difficulty of the proposed settings and compare them with the original setting that has real data and ground-truth labels. Our finding is that generally these proposed settings are more difficult than the original one, and are consequently useful to be applied on NAS. 
Furthermore, it should be noted that the feasibility of using unreal data unlock potentials for designing better NAS algorithms with the extra flexibility of designing data.

With this in mind, we conduct thorough evaluations on CIFAR-10, CIFAR-100, ImageNet and CheXpert~\cite{irvin2019chexpert} based on the standard differentiable architecture searching pipeline (e.g.~\cite{liu2018darts}).
Our UnrealNAS achieves 15.9\% test error on CIFAR-100 with 4.3M parameters. On ImageNet, UnrealNAS is able to find neural architectures with results (24.4\% Top-1 test error) comparable to state-of-the-arts.  
To further examine the generalization capability of the searched neural architectures, we apply UnrealNAS on medical dataset CheXpert (details in Appendix). We show that the real-data independent property of UnrealNAS is crucial, particularly in terms of finding appropriate neural architectures for scenarios with limited amount of available data. 

Our contributions are summarized as follows:
\begin{enumerate}
    \item We analyze the difficulty of proposed settings with unreal data and labels, showing the superiority of difficult tasks to better select neural architectures with higher effective capacity.
    \item We find counter-intuitive experimental results that UnrealNAS can achieve competitive performance without using real data or labels. Extensive experiments and ablation study validate the effectiveness of different unreal settings on selecting effective neural architectures. For instance, UnrealNAS with generated data achieves 24.4\% Top-1 test error on ImageNet using DARTS, while DARTS with original real training data and ground-truth labels can only get 26.7\% test error. 
    \item Data generation in UnrealNAS can be customized to strengthen the generalization ability of searched networks towards other tasks. It shows potentials for people to design better NAS algorithms together with the data they are trained on.
\end{enumerate}

%% file: Section/related.tex
\section{Related Work}
\label{sec:related}

\mysection{Neural Architecture Search}
\cite{zoph2016neural} made the first attempt to search neural architectures with reinforcement learning on hundreds of GPUs.
To reduce search cost, NASNet \cite{zoph2018learning}, ENAS \cite{pham2018efficient} and PNAS \cite{liu2018progressive} proposed techniques such as searching cell-based architectures, parameter sharing, and sequential model-based optimization strategy. 
However, thousands of GPU hours were still required to search one architecture. 
More recent one-shot/sampling-based NAS methods \cite{brock2017smash,bender2019understanding,guo2020single,ru2020neural,zela2018towards} have been proposed to reduce the search time by training a single over-parameterized supernetwork. 
DARTS \cite{liu2018darts} represented one important direction of research \cite{xie2018snas,wu2019fbnet,nayman2019xnas,zhou2019bayesnas,xu2019pc,chu2019fairdarts,li2019sgas,Zela2020Understanding,wang2021rethinking} that relaxed the discrete architecture selection in an end-to-end differentiable manner.
In this work, we examine another element of NAS efficiency and robustness by taking a closer look at differentiable NAS methods from a data-centric perspective. These findings can be beneficial for other NAS methods as well.

\mysection{Model Generalization and Capacity}
Recently, the generalization and capacity of DNNs have been actively studied to design more reliable network architectures~\cite{keskar2016large,cheng2018evaluating,lee2017ability,raghu2017expressive,novak2018sensitivity}. 
~\cite{zhang2016understanding} 
demonstrate CNNs trained with stochastic gradient methods can easily fit random labeled training data. 
\cite{neyshabur2017exploring} find simply accounting for complexity in terms of the number of parameters is not sufficient to explain the generalization capability of neural networks.
To investigate the model generalization and capacity, works~\cite{pondenkandath2018leveraging,maennel2020neural} study what neural networks learn when trained with random labels. In this work, we take one further step to explore what NAS algorithms would find if trained with unreal input data and random labels.

\mysection{NAS with Random/Self-Labeled Data}
UnNAS~\cite{liu2020labels} proposes to combine the self-supervised objective into DARTS so that ground-truth labels are not required. 
Another pioneering work is RLNAS~\cite{zhang2021neural} which takes original input data with only random labels for the single-path-one-shot searching (SPOS)~\cite{guo2020single}. 
Compared with UnNAS and RLNAS (as shown in Table~\ref{tab:comparison_nas}), UnrealNAS further explores the necessity of using real data. By applying a customized process of generating data with appropriate resolution and categories, UnrealNAS improves the generalization capability of searched networks to the target application. UnrealNAS can help better understand the mechanism behind NAS algorithms and explore the design space of dataset that NAS is performed on. 

\vspace{-5pt}
\begin{table}[ht]
    \centering
    \scalebox{0.9}{
    \begin{tabular}{lccccc}
         \toprule
         Method    &  Data      &  Label           & Selection Metric      \\
         \midrule
         DARTS     &  Original  &  Ground-Truth    &  Val Acc              \\
         \midrule
         UnNAS     &  Original  &  Self-Supervised &  Val Acc              \\
         \midrule
         RLNAS     &  Original  &  Random          &  Angle    \\
         \midrule
         \gc UnrealNAS &  Unreal    &  Random          &  Val Acc              \\
         \midrule
    \end{tabular}}
    \vspace{5pt}
    \caption{\textbf{Different settings in DARTS, UnNAS, RLNAS and UnrealNAS.}}
    \label{tab:comparison_nas}
\end{table}

%% file: Section/method_analysis.tex
\vspace{-5pt}

\section{Method and Analysis}
\label{sec:method}
\subsection{Motivation \& Analyses of NAS with Unreal Dataset}

To reduce the search cost, most NAS methods adopt a two-stage pipeline that searches for a good architecture on a proxy dataset, such as the CIFAR10 dataset, and then transfer the obtained architecture to a targeted dataset such as ImageNet, where the selected network is trained from scratch.
However, identifying the key to obtaining a good architecture using NAS on a proxy dataset is under-explored. First of all, do we really need a real dataset to identify a good architecture?
If a constructed dataset is enough to distinguish a good architecture from the poor ones, we may not need to collect any annotated dataset in order to find a good architecture. 
Inspired by \cite{zhang2017understanding} and \cite{KataokaACCV2020}, we create three kinds of unreal datasets which consist of only randomly labeled images, only generative images, and only generated Gaussian noise.

\begin{figure*}[ht]
\centering
\begin{minipage}{0.49\textwidth}
  \centering
  \includegraphics[width=0.99\textwidth]{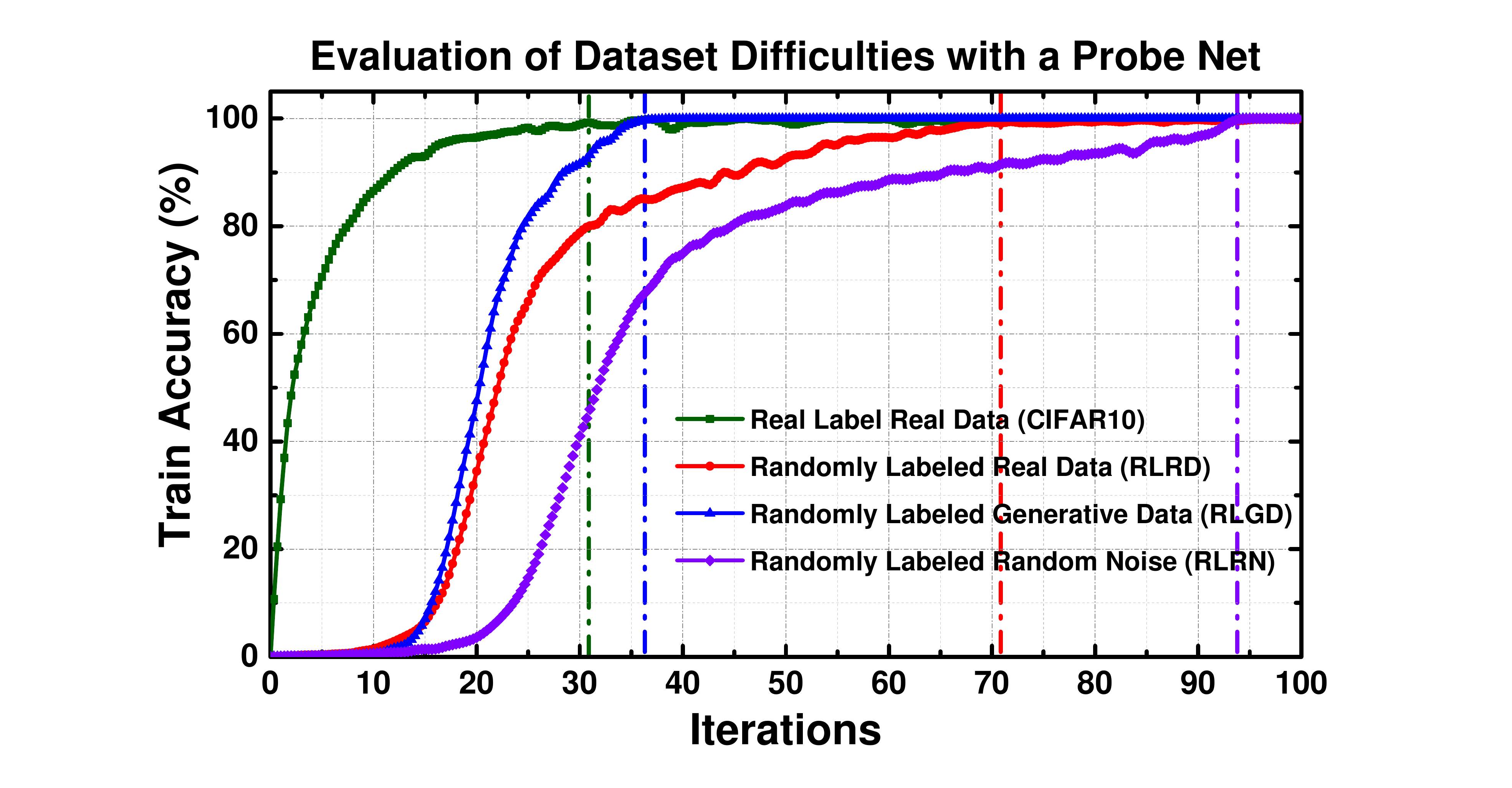} \\
    (a) Difficulties of Unreal Datasets.
    \label{fig:dataset_diff}
\end{minipage}
\begin{minipage}{0.49\textwidth}
  \centering
  \includegraphics[width=0.99\textwidth]{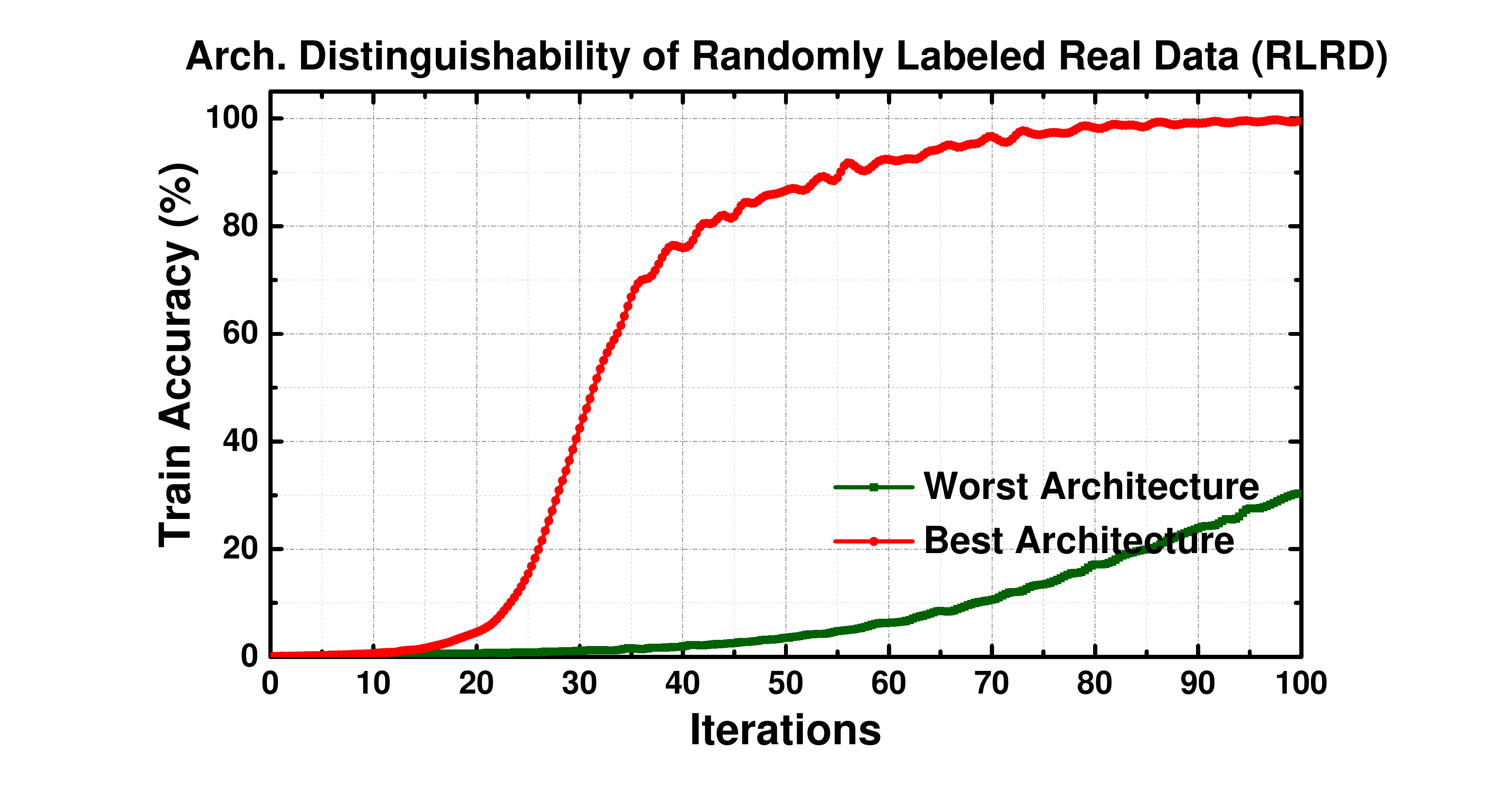} \\
    (b) Arch. Distinguishability of RLRD.
    \label{fig:dataset_diff}
\end{minipage}
\begin{minipage}{0.49\textwidth}
  \centering
  \includegraphics[width=0.99\textwidth]{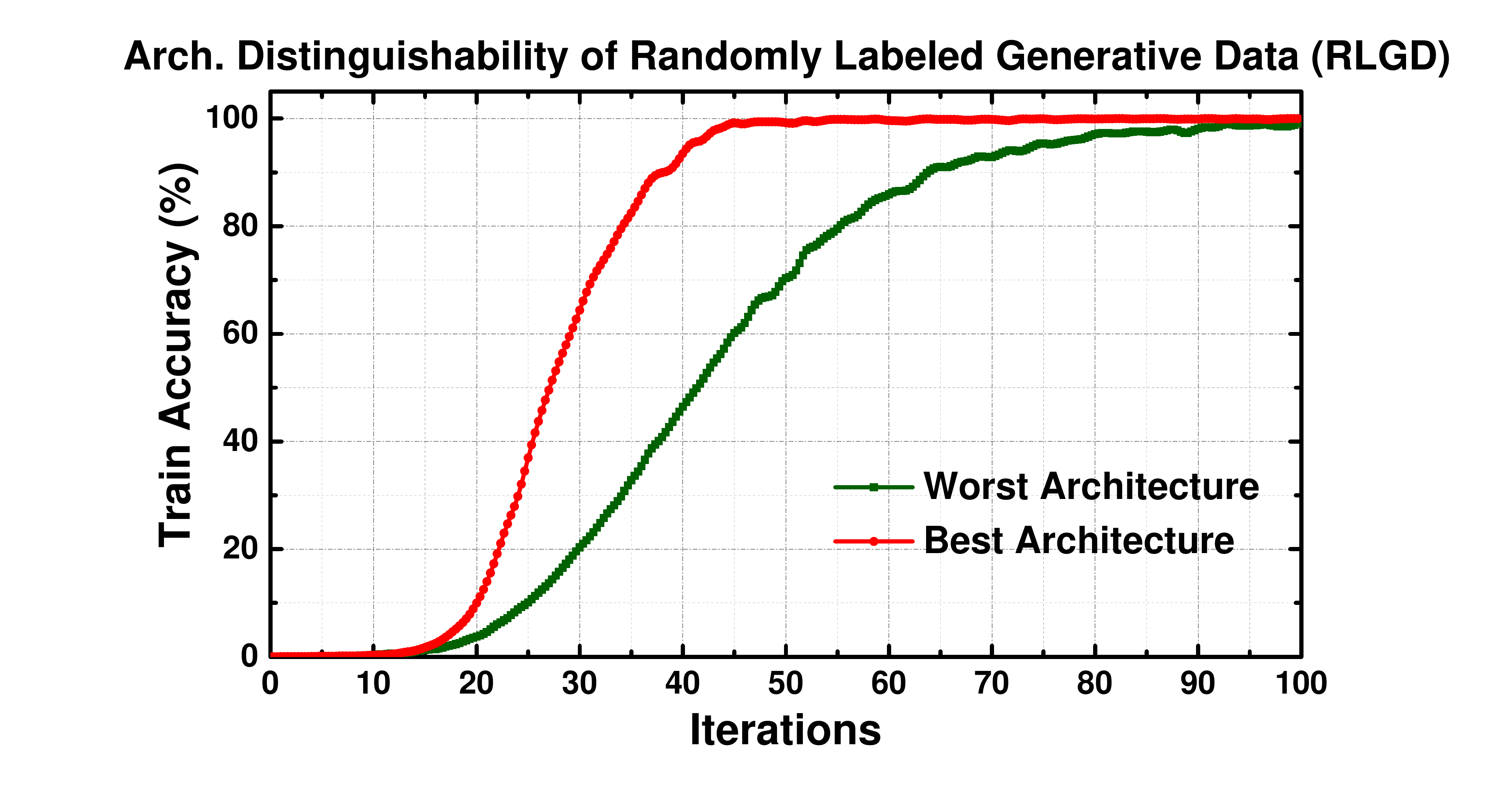} \\
    (c) Arch. Distinguishability of RLGD.
    \label{fig:dataset_dist}
\end{minipage}
\begin{minipage}{0.49\textwidth}
  \centering
  \includegraphics[width=0.99\textwidth]{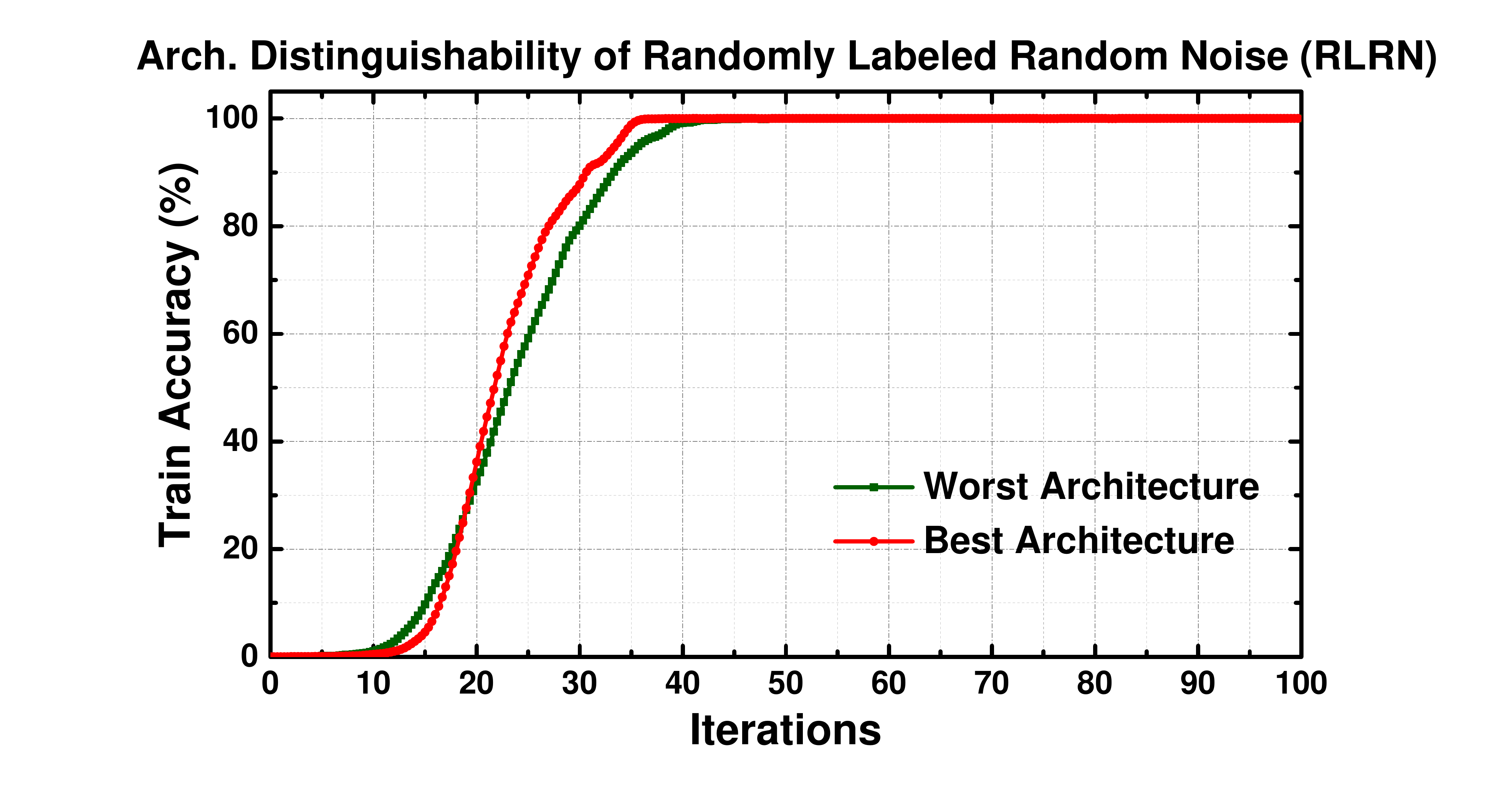} \\
    (d) Arch. Distinguishability of RLRN.
    \label{fig:dataset_dist}
\end{minipage}
\caption{\textbf{Analyses of Unreal Dataset Difficulties and Distinguishability.} (a) The convergence speeds of the probe net on different datasets show that the difficulties of unreal datasets are higher than CIFAR-10. (b-d) The learning curves of reference models show the architectural distinguishability of unreal datasets.}
\label{fig:analysis}
\end{figure*}

Intuitively, to identify a good architecture, the datasets should not be trivial to overfit by any architecture in the search space. To further investigate the characteristics of the unreal datasets, we conduct two empirical analyses: (1) evaluating the difficulties of unreal datasets using a randomly sampled architecture from the DARTS's search space as a probe net; and (2) evaluating the distinguishability of unreal datasets on the performance of architectures.  
In \figLabel \ref{fig:analysis} (a), we train the probe net from scratch on the unreal datasets and CIFAR-10. As can be seen, the randomly sampled probe net converges in different speed. Similar to \cite{scheidegger2020efficient}, we use the epoch to convergence as our scoring metric, where CIFAR-10, randomly labeled CIFAR-10, generative images and generated Gaussian noise score $31$, $71$, $36$ and $94$ respectively.
We observe that the probe net trained on CIFAR-10 (real data) converges fastest, and the training curve quickly reaches a plateau, which indicates the dataset is easy to overfit. 
The other constructed (unreal) datasets are shown to be more difficult to overfit than CIFAR-10.
In \figLabel \ref{fig:analysis} (b-d), we analyze the distinguishability of unreal datasets. We randomly sample two reference architectures from the DARTS search space and train them from scratch on CIFAR-100 to obtain the final test accuracies. Based on the accuracies, the two architectures are denoted as \emph{best} and \emph{worst}, respectively. We retrain the two reference models on unreal datasets and observe that the accuracies of reference architectures on unreal datasets at a probe (50-th) epoch are highly correlated to their accuracy ranking on CIFAR-100 (the \emph{best} model consistently outperforms the \emph{worst} model on unreal datasets). To further validate this finding, we randomly sample 80 architectures (20 on real dataset, RLRD, RLGD, RLRN respectively) and perform the same analysis. We calculate the Kendall Tau correlation ratio as a proxy to the correlation between rankings of CIFAR-100 accuracy and the 50th epoch training accuracy. The Kendall Tau correlation is 0.64 for RLRD, 0.60 for RLGD and 0.58 for RLRN, which suggests that training accuracy on unreal dataset has a strong correlation with CIFAR-100 accuracy. 
From these observations, we hypothesize that:
\begin{center}
\emph{If a dataset is difficult enough to distinguish architectures, it can be used for searching good architectures with a high effective capacity.} 
\end{center}

\subsection{Differentiable Neural Architecture Search} 
Cell-based NAS methods \cite{zoph2018learning,liu2018progressive,real2019regularized} learn scalable and transferable architectures by defining the search problem as finding the best cell structure. 
The architectures usually consist of repeating cell structures with different weights.
A cell can be represented as a directed acyclic-graph (DAG) with $N$ nodes,
where the $i$ topological ordered node $x^{(i)}$ denotes an intermediate feature representation.
Each directed edge $(i, j)$ is associated with an operation $o^{(i, j)}$ from the operation space $\mathcal{O}$.
DARTS \cite{liu2018darts} relaxes the discrete selection of operations to a continuous optimization problem by a softmax function.
Each edge is parameterized by architectural parameters $\alpha^{(i, j)}$ as a mixture operation of all the candidate operations within $\mathcal{O}$ during the search phase, \ie
$
	\bar{o}^{(i,j)}(x^{(i)}) = \sum_{o \in \mathcal{O}} \frac{\exp(\alpha_o^{(i,j)})}{\sum_{o' \in \mathcal{O}} \exp(\alpha_{o'}^{(i,j)})} o(x^{(i)})
$.
The feature map of each intermediate node is a summation of its predecessors,
$
	x^{(j)} = \sum_{i<j} \bar{o}^{(i, j)}(x^{(i)})
$.
The feature map of the output node is a concatenation of its several predecessors. Let $\mathcal{W}$ denotes the weights of the supernet and $\mathcal{A}$ denotes the architectural parameters. $\mathcal{A}$ and $\mathcal{W}$ are jointly learned via a bi-level optimization:
\begin{align}
    	\min_\mathcal{A} \quad  \mathcal{L}_{val}(\mathcal{W}^*(\mathcal{A}), \mathcal{A}) \quad ~~~~~~~~\\
    	\text{s.t.} \quad \mathcal{W}^*(\mathcal{A}) = \mathrm{argmin}_\mathcal{W} \enskip \mathcal{L}_{train}(\mathcal{W}, \mathcal{A})
    	\label{eq:bilevel}
\end{align}
$\mathcal{L}_{train}$ and $\mathcal{L}_{val}$ denote the losses on the training set $\mathcal{D}_{train}$ and validation set $\mathcal{D}_{val}$ respectively. The final cell is derived by selecting the operation with highest weight from each mixture operation,
$
	o^{(i,j)} = 
	\mathrm{argmax}_{o \in \mathcal{O}} \enskip \alpha^{(i,j)}_o
$.

\begin{figure*}[t]
\centering
\includegraphics[width=0.99\textwidth]{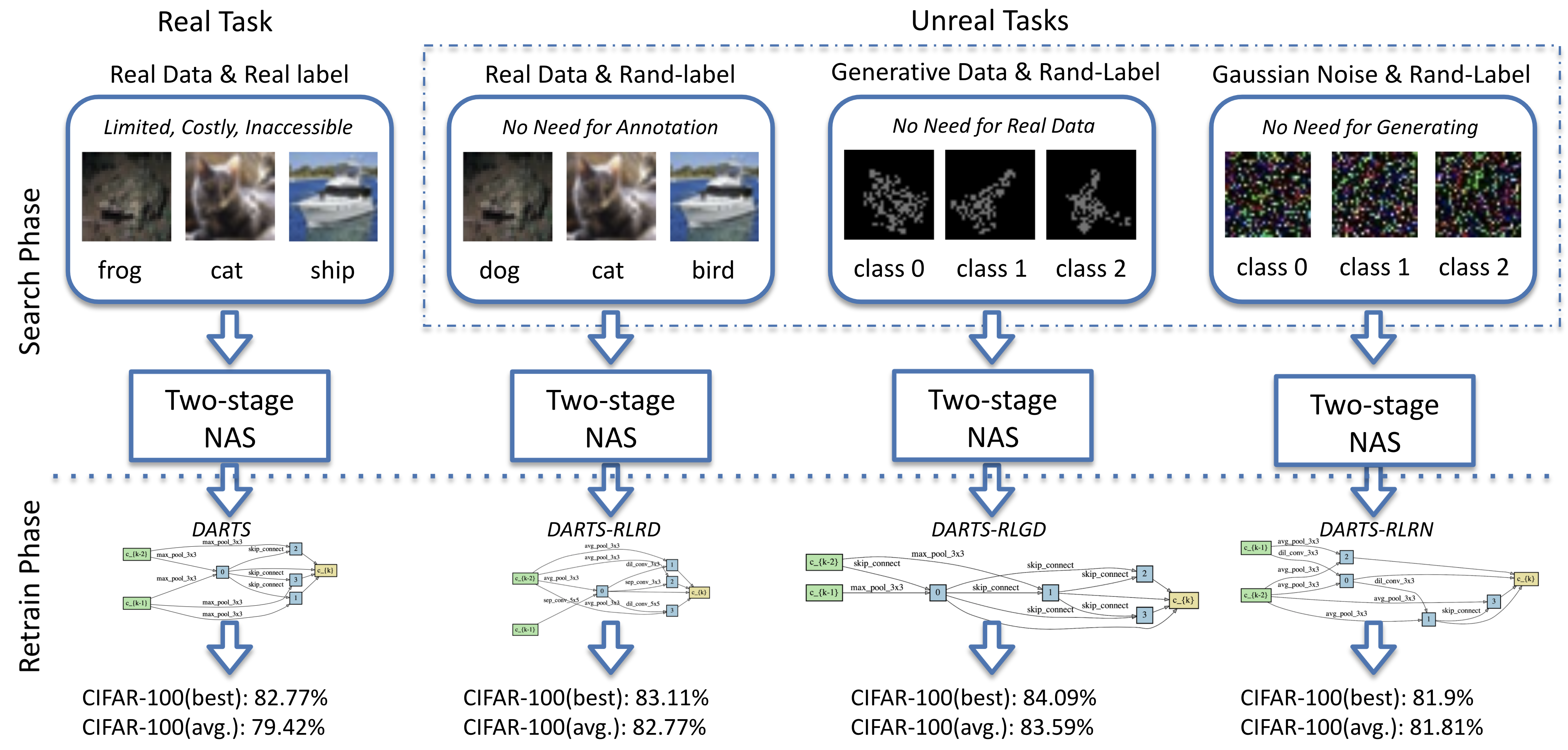}
\caption{
\textbf{UnrealNAS \vs Conventional Two-Stage NAS Methods.} UnrealNAS keeps all the settings identical to the original NAS method except for using unreal datasets for searching architectures. CIFAR-100 best and average performances are reported with 5 repetitions. The best result of DARTS is obtained by retraining DARTS-V1 (\cite{liu2018darts}). Average result is obtained from (\cite{Zela2020Understanding}) where they adopt identical setting and 5 repetitions. By simply changing searching datasets, UnrealNAS obtains substantial improvement over original methods.
}
\label{fig:pipeline}
\end{figure*}

\subsection{UnrealNAS} 

In this work, 
we aim to perform NAS without real annotations or real images. We explore three different ways to construct unreal datasets for image classification. 
In the following section, we first discuss the generating process of three types of unreal datasets with the training set $\mathcal{D}_{train}$ containing $n$ distinct data points $\set{(x_i, y_i)}$, where $x_i$ are $H\times W \times C$ images and $y_i$ are $d$-dimensional one-hot labels. We then explain how to search neural architectures with these unreal datasets.

\begin{figure}[t]
\centering
\scalebox{0.7}{
\includegraphics[scale=0.3]{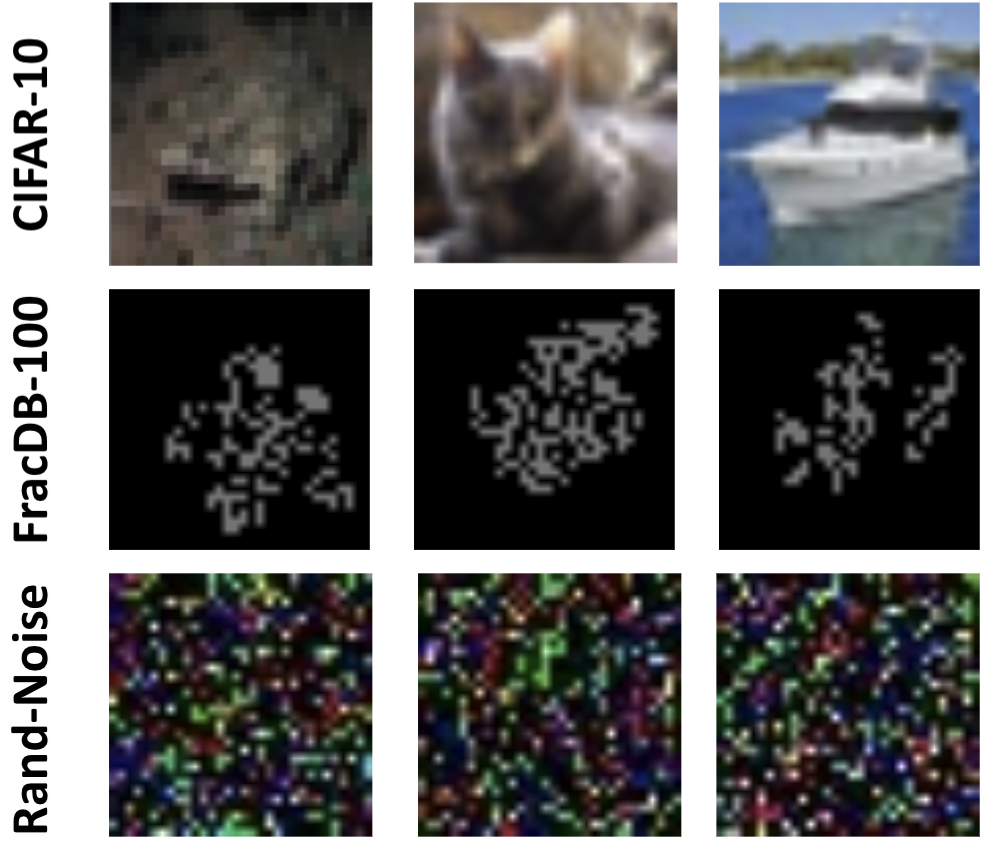}}
\qquad
\scalebox{0.8}{
\begin{tabular}[b]{lccccc}
\hline
\multicolumn{1}{c}{\begin{tabular}[c]{@{}c@{}}Dataset\\ Name\end{tabular}} & \begin{tabular}[c]{@{}c@{}}Dataset\\ Source\end{tabular} & \begin{tabular}[c]{@{}c@{}}Dataset \\ Size\end{tabular} & $H\times W$ & \#ch & \#cls \\
\midrule
RLRD & \begin{tabular}[c]{@{}c@{}}Real World\\ (CIFAR10)\end{tabular} & \begin{tabular}[c]{@{}c@{}}train: 25k\\ val: 25k\end{tabular} & $32 \times 32$ & 3 & 1k \\
\midrule
RLGD & \begin{tabular}[c]{@{}c@{}}Generative\\ (IFS)\end{tabular} & \begin{tabular}[c]{@{}c@{}}train: 25k\\ val: 25k\end{tabular} & $32 \times 32$ & 3 & 5k \\
\midrule
RLRN & \begin{tabular}[c]{@{}c@{}}Normal Distribution\\ ($\mu=0$, $\sigma=1$)\end{tabular} & \begin{tabular}[c]{@{}c@{}}train: 25k\\ val: 25k\end{tabular} & $32 \times 32$ & 3 & 5k \\ \midrule
\end{tabular}}
\caption{\textbf{Illustrations and statistics of unreal datasets.} Following the standard data split of DARTS, one half of the 50k total samples are used for training and the other half are used for validation. For random labeled classes, 1k classes are assigned to CIFAR-10 and 5k random classes are assigned to the random noise and generated data.}
\label{fig:unreal_datasets}
\end{figure}

\mysection{Randomly Labeled Real Data (RLRD)} Following DARTS, we randomly take $n=25K$ images $\set{(x_i, y_i)}$ from the training set of CIFAR-10 as the training images. Instead of using the true labels $y_i$, we randomly generate $d^{rand}$-dimensional one-hot labels for $x_i$. The randomly labeled datasets can easily achieve low Silhouettes scores \cite{rousseeuw1987silhouettes}, which means they are harder to classify. Since the images can be labeled in an arbitrary number of classes, we also investigate different numbers of classes with $d^{rand} \in \set{2,  50, 100, 200, 1000, 2000, 5000, 10000}$. In practice, we generate the one-hot $y_i^{rand}$ label for $x_i$ with a discrete uniform distribution $\Pr(y_{ij}^{rand}=1|\sum_{j}y_{ij}^{rand}=1) = 1/d^{rand}$. In this way, we construct a randomly labeled CIFAR dataset $\set{(x_i^{real}, y_i^{rand})}$.

\mysection{Randomly Labeled Generative Data (RLGD)} Inspired by the FractalDB dataset \cite{KataokaACCV2020}, we generate a formula-driven dataset with fractal images based on the iterated function system (IFS).
We assigned 100 [category] $\times$ 600 [instance] as a basic dataset size. The categories are associated with a set of six parameters i = $(a_i, b_i, c_i, d_i, e_i, f_i)$ for rotation and shifting. Intra-category images are generated by changing one of the parameters $a_i, b_i, c_i, d_i, $ and $e_i, f_i $ with weighting parameter $ w$. Following the setting of FractalDB,  we apply a 3×3 patch filter to generate fractal images in addition to the rendering at each 1×1 point with the filling rate of 0.2.
The generated fractal images have the same size ($32\times32\times3$) as CIFAR-10.
The labels of fractal images are also randomly assigned as the randomly labeled CIFAR dataset with $d^{rand}=\set{2, 50, 100, 200, 1000, 2000, 5000, 10000}$ dimensions. As a result, we construct a randomly labeled fractal dataset $\set{(x_i^{frac}, y_i^{rand})}$.

\mysection{Randomly Labeled Random Noise (RLRN)} For the random noise data, $x_i$ are randomly generated tensors with a resolution of $32\times32$ and a channel size of $C=3$. The value of each pixel and channel in $x_i$ is sampled independently from a standard normal distribution $\mathcal{N}(0,1)$. The labels of random noise images are also randomly assigned based on a uniform distribution with a labeling dimension $d^{rand}=\set{10, 100, 1000, 5000, 7500, 10000}$. We denote the RLRN dataset as $\set{(x_i^{noise}, y_i^{rand})}$.

\mysection{Search architectures with Unreal Data} To perform the bi-level optimization (\eqnLabel \ref{eq:bilevel}) in DARTS, the validation sets of unreal datasets are constructed from the training sets $\set{(x_i^{real}, y_i^{rand})}$, $\set{(x_i^{frac}, y_i^{rand})}$ and \\ $\set{(x_i^{noise}, y_i^{rand})}$ respectively by transforming the input images $x_i$ to $x^{\prime}_i$ through a standard data augmentation while leaving the labels unchanged. 

We show some illustrations and the statistics for each unreal dataset in \figLabel \ref{fig:unreal_datasets}. We compare UnrealNAS with the conventional two-stage NAS pipeline in \figLabel \ref{fig:pipeline}. The only modification needs to be made to conduct UnrealNAS from existing NAS methods is to change the dataset loader, which makes UnrealNAS easy to be applied for different searching algorithms. Due to the flexibility of manipulating our unreal dataset, different settings such as dataset size, the input resolution, the number of categories, and the prediction tasks can be adapted for the targeted downstream tasks with minimum effort. This exhibits the potential of UnrealNAS in searching architectures for various tasks in many real world applications.

%% file: Section/experiment.tex
\section{Experiments}\label{sec:experiment}
We evaluate three tasks on unreal data or labels, which can be used to perform differentiable architecture search, and find even better architectures than searching on real-world data with ground-truth labels. The experiments are primarily conducted on DARTS search space. Since recent works~\cite{yang2019evaluation,li2020random,yu2019evaluating} question the distinguishability of architecture searching on DARTS search space, we also apply UnrealNAS on top of more advanced NAS algorithm SGAS~\cite{li2019sgas}. On the search phase, we replace the original dataset with 1) CIFAR-10 data with random labels, 2) FracDB-100 generated data with random labels and 3) random noise with random labels. On the retrain phase, we retrain the selected architectures on CIFAR-100, ImageNet and CheXerpt (a medical dataset), comparing results with architectures searched on real-world data.

\subsection{Experimental Setup}
\mysection{Dataset}
In the search phase, we experiment on three types of data: CIFAR-10, FracDB-100 (generated data) and random noise. CIFAR-10 consists of real-world images with 10 classes. It has 50,000 images for training and 10,000 for validation.
FracDB-100 is a dataset generated using iterative fractal function. It has 100 classes and each label is denoted using the set of initial parameters. The resolution of FracDB-100 is equivalent to CIFAR-10, namely $32 \times 32$ with 3 channels. Random noise data are generated by pure Gaussian noise with each pixel belongs to Gaussian distribution (mean = 0 and std = 1). The size of random noise dataset and the data resolution are equivalent to CIFAR-10, with the only difference that the random pixels are in floating point values instead of unsigned int8 gray scales.
In the retrain phase, we use CIFAR-10, CIFAR-100, ImageNet and CheXpert to validate the generalization performance of our selected architectures. CIFAR-100 consists of 50,000 training images and 10,000 validation images with 100 classes. ImageNet (ILSVRC2012) consists of over 1.2M training images and 50K images for validation. 
CheXpert contains chest x-ray images and is used to diagnose 5 different thoracic pathologies: atelectasis, cardiomegaly, consolidation, edema and pleural effusion (more information in Appendix).

\mysection{Unreal Task Formulation}
We replace ground-truth in the datasets with random labels, where the label of each class is assigned with uniform probability.
Since we use random labels, the number of classes is adaptive in UnrealNAS. In our experiments, we use randomly labeled 1000 classes for CIFAR-10, 5000 classes for FracDB-100, and 5000 classes for random noise setting. In the search phase, we split the original training set in half to form the actual training set and validation set, following the same setting in DARTS. For the training set, we only apply normalization as preprocessing, while for the validation set, we use the standard horizontal flip as an augmentation. This is helpful to differ the data distribution of training and validation set, so that the validation accuracy can better evaluate the effective capacity of searched networks by evaluating both their memorization and generalization ability.

\mysection{Implementation Details}
In our experiments, we search 55 epochs in total with 5 warm-up epochs. To enable the ability of fitting unreal task, we set arch weight decay and weight decay all as 0. Since we only focus on the input data, all hyperparameters are kept identical to DARTS (more details in the Appendix).

\begin{table*}[t]
\centering
\resizebox{0.99\linewidth}{!}{%
\begin{tabular}{lcccccc}
\toprule
\textbf{{Architecture}}  & \multicolumn{2}{c}{\textbf{Test Err. (\%)}} & \textbf{Params}  & \textbf{Search Cost} &\textbf{Search}&\textbf{{Search}}  \\
&                            \textbf{CIFAR10} &\textbf{CIFAR100} & \textbf{(M)} &  \textbf{(GPU-days)} & \textbf{Data} &\textbf{Label} \\
\midrule
NASNet-A \cite{zoph2018learning} & 2.65 & - & 3.3 & 1800 & CIFAR10 & Real Label \\ 
AmoebaNet-B \cite{zoph2018learning} & 2.55$\pm$0.05 & - & 2.8 & 3150 & CIFAR10 & Real Label  \\ 
PNAS \cite{zoph2018learning} & 3.41$\pm$0.09 & - & 3.2 & 225 & CIFAR10 & Real Label  \\ 
ENAS \cite{real2019regularized} & 2.89 & - & 3.3 & 0.5 & CIFAR10 & Real Label \\ 
NAONet \cite{real2019regularized} & 3.18 & 15.67 & 10.6 & 200 & CIFAR10 & Real Label \\ 
\midrule
DARTS \cite{Zela2020Understanding} & 2.91$\pm$0.25 & 20.58$\pm$0.44 & 3.3 & 1.5 & CIFAR10 & Real Label\\ 
DARTS \cite{liu2018darts} & 3.0$\pm$0.14  & 17.23 & 3.3 & 1.5 & CIFAR10 & Real Label\\ 
SNAS (moderate)  \cite{xie2018snas} & 2.85 & - & 2.8 & 1.5 & CIFAR10 & Real Label \\ 
ProxylessNAS \cite{cai2018proxylessnas} & 2.08 & - & 5.7 & 4 & CIFAR10 & Real Label \\ 
P-DARTS \cite{chen2019progressive} & 2.50 & 16.55 & 3.4 & 0.3 & CIFAR10 & Real Label \\ 
P-DARTS \cite{chen2019progressive} & 2.62 & 15.92 & 3.6 & - & CIFAR100 & Real Label \\ 
ASAP \cite{zhou2019bayesnas} & 2.49$\pm$0.04 & 15.6 & 2.5 & 0.2 & CIFAR10 & Real Label  \\ 
PC-DARTS \cite{xu2019pc} & 2.57$\pm$0.07 & - & 3.6 & 0.1 & CIFAR10 & Real Label \\
FairDARTS \cite{chu2019fairdarts} &2.54$\pm$0.05 & - & 3.32$\pm$0.46 &- & CIFAR10 & Real Label\\
R-DARTS(L2)  \cite{Zela2020Understanding} & 2.95 $\pm$0.21 & 18.01 $\pm$0.26 & - & - & CIFAR10 & Real Label\\
SGAS (Cri.2)\cite{li2019sgas}& 2.67$\pm$0.21 & - & 4.1 & 0.25  & CIFAR10 & Real Label \\
\midrule
\gc UnrealNAS-DARTS-RLRD & 2.62(2.71$\pm$0.08) & 16.89(17.23$\pm$0.24) & 3.0 & 1.5 & CIFAR10 & Random Label \\
\gc UnrealNAS-SGAS-RLGD & 2.44(2.60$\pm$0.11) & 15.75(16.29$\pm$0.26) & 4.0 & 0.25 & FracDB-100(generative) & Random Label \\
\gc UnrealNAS-DARTS-RLGD & 2.46(2.62$\pm$0.10) & 15.91(16.42$\pm$0.27) & 4.3 & 1.5 & FracDB-100(generative) & Random Label  \\
\gc UnrealNAS-DARTS-RLRN & 2.61(2.73$\pm$0.12) & 17.26(17.54$\pm$0.24) & 3.0 & 1.5 & Gaussian Noise & Random Label  \\
\bottomrule
\end{tabular}}
\vspace{5pt}
\caption{\textbf{Results with different architectures on CIFAR10/100.} Both the average and best results of UnrealNAS are reported.}
\label{tab:compare_cifar}
\end{table*}

\subsection{Results}

\begin{table*}[t]
\centering
\resizebox{0.99\linewidth}{!}{%
\begin{tabular}{lcccccc}
\toprule
\textbf{{Architecture}}&\multicolumn{2}{c}{\textbf{Test Err. (\%)}}&\textbf{Params}& \textbf{Search Cost} & \textbf{Search} & \textbf{{Search}} \\
& \textbf{top-1} & \textbf{top-5} & \textbf{(M)} & \textbf{(GPU-days)} & \textbf{Data} & \textbf{Label} \\
\midrule
NASNet-A \cite{zoph2018learning} & 26 & 8.4 & 5.3 & 1800 & CIFAR10 & Real Label \\ 
AmoebaNet-A \cite{real2019regularized} & 25.5 & 8 & 5.1  & 3150 & CIFAR10 & Real Label \\ 
FairNAS-A \cite{chu2019fairdarts} & 24.7 & 7.6 & 4.6 & 12 & CIFAR10 & Real Label \\
PNAS \cite{liu2018progressive} & 25.8 & 8.1 & 5.1 & 225 & CIFAR10 & Real Label \\ 
MnasNet-92 \cite{tan2019mnasnet} & 25.2 & 8 & 4.4 & - & CIFAR10 & Real Label \\
\midrule
DARTS \cite{liu2018darts} & 26.7 & 8.7 & 4.7 & 4.0 & CIFAR10 & Real Label\\ 
SNAS (mild) \cite{xie2018snas} & 27.3 & 9.2 & 4.3 & 1.5 & CIFAR10 & Real Label \\ 
ProxylessNAS \cite{cai2018proxylessnas} & 24.9 & 7.5 & 7.1  & 8.3 & ImageNet & Real Label \\ 
P-DARTS \cite{chen2019progressive} & 24.4 & 7.4 & 4.9 & 0.3 & CIFAR10 & Real Label \\ 
BayesNAS \cite{zhou2019bayesnas} & 26.5 & 8.9 & 3.9 & 0.2 & CIFAR10 & Real Label \\ 
PC-DARTS \cite{xu2019pc} & 25.1 & 7.8 & 5.3 & 0.1 & CIFAR10 & Real Label \\
SemiNAS~\cite{luo2020semi}   & 23.5 & 6.8 & 6.3 & 4  & ImageNet & Real Label \\
SGAS (Cri.1)\cite{li2019sgas}& 24.4 & 7.3 & 5.3 & 0.25  & CIFAR10 & Real Label \\
\midrule
UnNAS~\cite{liu2020labels}   & 24.1 & -   & 5.2 & 2  & ImageNet & Self-Supervised Label \\
RLNAS~\cite{zhang2021neural} & 24.4 & 7.5 & 5.3 & -  & CIFAR-10 & Random Label \\
RLNAS~\cite{zhang2021neural} & 24.1 & 7.1 & 5.5 & -  & ImageNet & Random Label \\
\midrule
\gc UnrealNAS-DARTS-RLRD & 25.8 & 8.1 & 4.4 & 4.0 & CIFAR10 & Random Label \\
\gc UnrealNAS-SGAS-RLGD & 24.4 & 7.5 & 6.4 & 0.25 & FracDB-100(generative) & Random Label \\
\gc UnrealNAS-DARTS-RLGD & 24.4 & 7.6 & 5.8 & 4.0 & FracDB-100(generative) & Random Label  \\
\gc UnrealNAS-DARTS-RLRN & 25.4 & 8.1 & 4.7 & 4.0 & Gaussian Noise & Random Label  \\
\bottomrule
\end{tabular}}
\vspace{3pt}
\caption{\textbf{Comparison with state-of-the-art NAS methods on ImageNet.} We transfer the top $2$ architectures searched on unreal dataset to ImageNet.}
\savespace
\label{tab:compare_imagenet_performance}
\end{table*}

\mysection{CIFAR10/100}
As shown in Table~\ref{tab:compare_cifar}, for CIFAR-100, UnrealNAS with generated data achieves 15.91\% test error using DARTS, while DARTS with original training data and ground-truth labels can only get 17.23\% test error. In addition to the best performance, we also report the average test error 16.42\% with standard deviation 0.27\%, which consistently outperforms the original DARTS while having a smaller parameter size. Furthermore, to show the effectiveness of UnrealNAS, we compare it with other advanced NAS algorithms such as P-DARTS~\cite{chen2019progressive}, FairDARTS~\cite{chu2019fairdarts} and R-DARTS~\cite{Zela2020Understanding}. As we can see, UnrealNAS can obtain very competitive results compared to the state-of-the-art NAS algorithms, with only generated data and random labels. When using real CIFAR-10 data with random labels, the final results on CIFAR-100 become 0.8\% worse than generated data, but with a smaller parameter size (3.0M compared to 4.3M). For the setting with randomly labeled random noise, the final results on CIFAR-100 become 0.3\% worse than randomly labeled real data, while still being comparable to other NAS methods using real data and labels. We find similar trend for UnrealNAS and state-of-the-art methods on CIFAR-10 results. In order to show the effectiveness of UnrealNAS on top of other method and space, we also provide the results on SGAS. As can be seen, our UnrealNAS-SGAS results outperform SGAS on both accuracy and parameter efficiency (more results in Table~\ref{tbl:compare_different_search_methods} in Appendix).

\mysection{ImageNet}
In Table~\ref{tab:compare_imagenet_performance}, we compare UnrealNAS results on ImageNet with networks searched by state-of-the-art NAS methods. As can be seen, UnrealNAS with CIFAR-10 data and random labels achieves 25.8\% test error, which outperforms the 26.7\% of DARTS by a large margin. Using randomly labeled random noise can yield similar results as CIFAR-10 data. When taking generated data and labels as inputs, UnrealNAS is able to achieve 24.4\% test error, which surpasses most of the state-of-the-art results while being comparable to strong baselines like BayesNAS~\cite{zhou2019bayesnas}. UnNAS~\cite{liu2020labels} achieves 24.1\% test error with self-supervised label (Jigsaw Puzzles), which outperforms UnrealNAS by 0.3\%. However, it should be noted that UnNAS is directly performed on ImageNet data (unlike other state-of-the-art NAS that are on CIFAR-10). The same trend applies for RLNAS~\cite{zhang2021neural} where UnrealNAS achieves comparable final accuracy without access to original data. Based on these comparisons, we can conclude that unreal data with random labels are capable to select decent network architectures on ImageNet.

\mysection{CheXpert}
In Table~\ref{tbl:CheXpert}, we compare the performance of state-of-the-art neural network models on CheXpert(results from~\cite{ke2021chextransfer}) with our UnrealNAS results. We follow the same training strategy as~\cite{ke2021chextransfer}. It can be seen that the network architecture found by generated data can achieve better accuracy (with or without pretraining on ImageNet) while achieving smaller parameter size. This indicates the effectiveness of our UnrealNAS searched architectures on different image domain and on different tasks.

\begin{table}[h]
\centering
\resizebox{0.8\linewidth}{!}{%
\begin{tabular}{lcccccccc}
\toprule
\textbf{{Architecture}}  & \textbf{Params (M)} & \textbf{w/o pretrain (\%)} & \textbf{w/ pretrain (\%)} \\
\midrule
EfficientNetB0 & 4.03         & 83.6 & 85.9 \\ 
EfficientNetB1 & 6.53         & 83.8 & 85.8 \\
MobileNetV3    & 4.22         & 82.7 & 85.9 \\
ResNet18       & 11.7         & 84.5 & 86.2 \\
ResNet50       & 25.6         & 85.4 & 85.9 \\
\midrule
\gc UnrealNAS-DARTS-RLRD & 3.17 & 83.8 & 85.0  \\
\gc UnrealNAS-DARTS-RLGD & 4.32 & 85.4 & 86.1  \\
\gc UnrealNAS-DARTS-RLRN & 2.99 & 83.8 & 85.2  \\
\bottomrule
\end{tabular}}
\vspace{10pt}
\caption{\textbf{Comparison with SOTA Models on CheXpert.}}
\label{tbl:CheXpert}
\end{table}

%% file: Section/ablation.tex
\section{Ablation Study}
\label{sec:ablation_study}
\vspace{-5pt}

\subsection{Selected Cells and Search Tasks Discussion}
\begin{figure}[ht]
\centering
\scalebox{1}{
\includegraphics[width=0.99\textwidth]{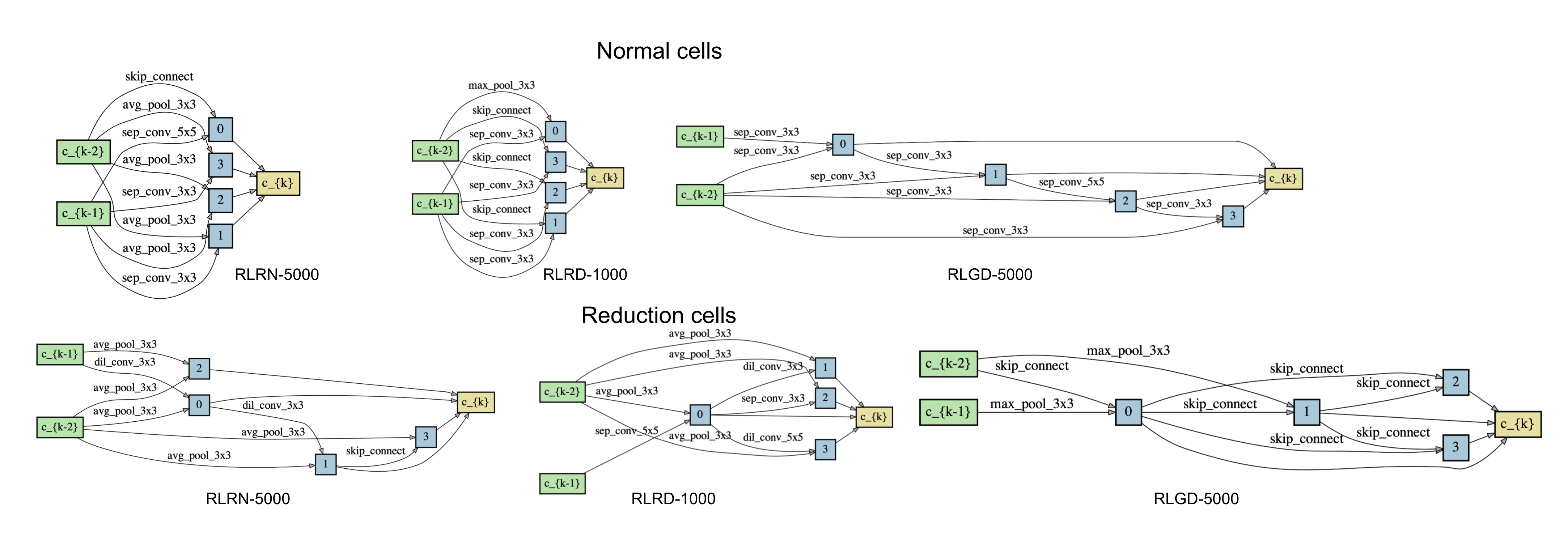}}
\caption{\textbf{Cell of best structures searched by UnrealNAS.} The normal cells and reduce cells are shown. The obtained cells searched by UnrealNAS have reasonable capacity and receptive fields. Please zoom for better view.}
\label{fig:cells}
\end{figure}

The experiment results above is counter-intuitive because the search tasks are unreal but can lead to decent architecture. We visualize the obtained cells in \figLabel \ref{fig:cells}, where we observe that the cells searched on unreal tasks actually contain operators with reasonable effective capacity and receipt fields. In addition, the overall structure of searched cells are non-shallow. The decency of cells searched with our random setting suggests that we may need to rethink the mechanism behind prevailing NAS algorithms.

\vspace{-15pt}
\subsection{Analysis of Skip Connections Collapsing}
\begin{figure}[ht]
\centering
  \includegraphics[trim=140 20 130 20, clip, width=0.6\textwidth]{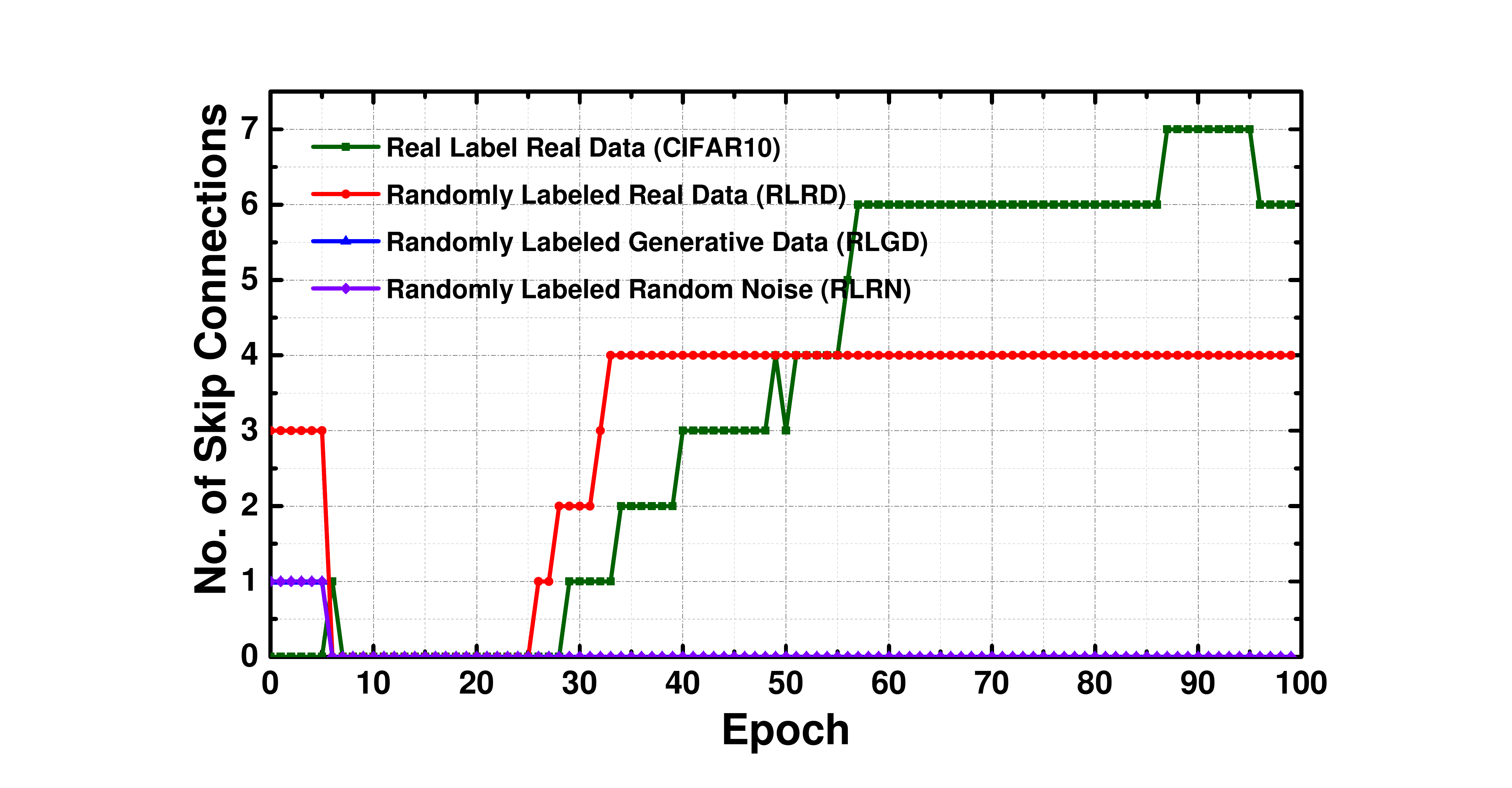} \\
    \caption{\textbf{Num of skip-connections. $\vs$ Search epochs.} The traditional DARTS (search on real data) demonstrates the issue of collapse in skip-connection operators. For UnrealNAS-DARTS, there is no sign of collapse even till 100 epochs.}
    \label{fig:dynamic}
\end{figure}

We show the search dynamics analysis in \figLabel \ref{fig:dynamic}. As can be seen, DARTS on real data demonstrates serious collapse issue. When the search epochs reach over 57, the number of skip connections is exploding to 6, while typically architecture with too many skip-connections tend to become shallow with low effective capacity. In contrast, DARTS on unreal datasets has alleviated the collapsing issue, where we run 100 search epochs and find that the number of skip connections is still within a reasonable range.

\subsection{Impact of randomly labeled classes}
\begin{figure}[h]
  \centering
  \includegraphics[trim=120 0 140 20, clip, width=0.6\textwidth]{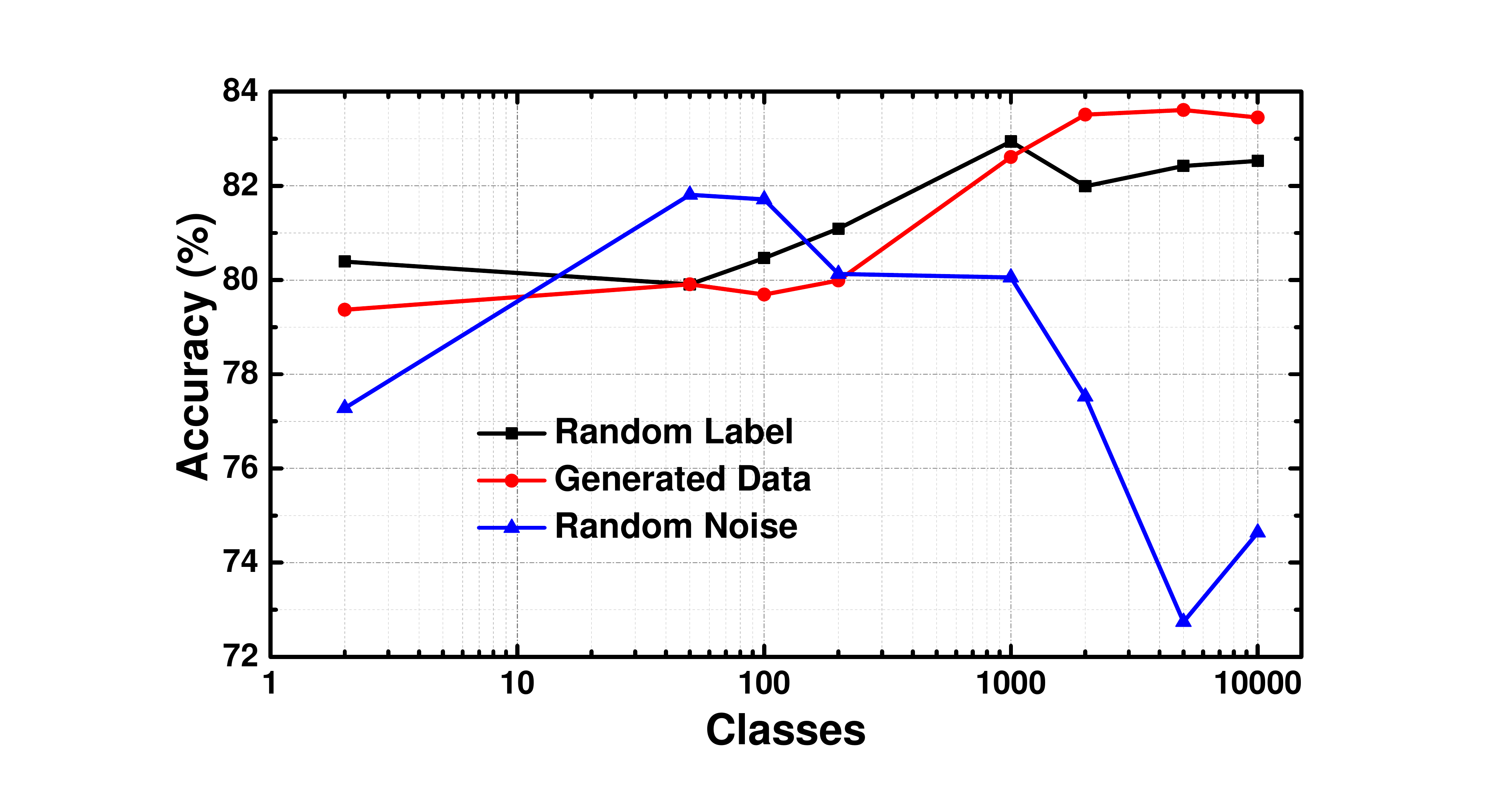} \\
    \caption{\textbf{Impact of different random labeled number of classes on test accuracy on CIFAR-100.} We ablate the number of classes from 2 to 10000. The search performance gains with a properly larger number of classes.}
    \label{fig:num_lasses}
\label{fig:exp_analysis}
\end{figure}

As \figLabel \ref{fig:num_lasses} shows, the number of random label classes have considerable impact on selecting good architectures. The number of random label classes is related to the task difficulty. When data per classes is sufficient, more random classes will positively help select architectures with high effective capacity (Randomly Labeled CIFAR and Generated Data with a class number from 2-1000). When the data per class is too few, more classes will negatively influence the selection method and even cause some instability of searching performance (Random Noise with class number over 100).

%% file: Section/conlusion.tex
\section{Conclusion}
A key element of NAS is the data used to evaluate the searched architectures. 
We explore whether NAS can be conducted with unreal data by proposing UnrealNAS.
We analyze the difficulty of proposed settings with unreal data and labels, showing the superiority of difficult tasks to better select neural architectures with higher effective capacity.
Extensive experiments and ablation study validate the effectiveness of different settings on selecting effective neural architectures. 
UnrealNAS achieves competitive results without using real data or labels. 
Aside from the fundamental importance of investigating the role of data in NAS, the use of unreal data has important practical impact, for the data or annotations may not be available in many applications due to privacy or cost constraint.

%% file: Section/supp.tex
\section{Implementation Details}\label{supp:additionalexperiment}
\subsection{Searching Phase}
\mysection{Data Pipeline}
We load half of the training data as train set. For Real Label Real Data setting, the other half of training is used as validation data. The data augmentation for train set is composed of random crop, horizotal flip (p=0.5) and normalization of CIFAR10 mean/std. The data augmentation for validation set is only normalization. 
For all unreal datasets, validation set is using the same half training data as train set. We only apply normalization on the train set. To maintain certain correlation with the train set, while also having some divergence from the train set, we apply horizontal flip and normalization on the validation set.

\mysection{Search Hyperparameter}
We train a tiny network with 8 cells (normal cell and reduction cell) for 5 warm-up epochs and 50 searching epochs with batch size of 64. We mainly search on single GPU. We use momentum SGD as optimizer (momentum = 0.9, weight decay = 0 for unreal tasks, weight decay = 0.0003 for real tasks). The initial learning rate is 0.025 with cosine schedule annealing to zero without restart. To ensure equal attention for all operators, zero initialization is used to initialize architecture variables. For updating architecture weights, Adam is set as optimizer with initial learning rate = 0.0003, momentum = (0.5, 0.99) and arch weight decay = 0 for unreal tasks, 0.001 for real tasks.

\subsection{Retraining Phase}
\mysection{CIFAR10/100}
We train 500 epochs for a large network of 20 cells with 36 channels. The batch size is 96 and the training is on one single GPU.

\mysection{ImageNet}
We use input resolution $224\times 224$. The whole network consists of 14 cells. To make retraining time manageable, we train it for 200 epochs with 1024 batch size on 8 GPU. We use an initial SGD learning rate of 0.5 decayed by a factor of 0.97 for each epoch.

\section{Medical data - CheXpert}\label{supp:medical}
CheXpert~\cite{irvin2019chexpert} is a large dataset of chest X-rays and competition for automated chest x-ray interpretation, which features uncertainty labels and radiologist-labeled reference standard evaluation sets.
CheXpert contains 224,316 chest radiographs of 65,240 patients. It captures uncertainties labels of 14 observations in radiograph interpretation. The validation set of CheXpert contains 200 chest radiographic studies which were manually annotated by 3 board-certified radiologists. And the test set is composed of 500 chest radiographic studies annotated by a consensus of 5 board-certified radiologists (atelectasis, cardiomegaly, consolidation, edema and pleural effusion).

\section{Algorithm Agnostic}\label{supp:additionalexperiment}
To show the generalization ability of UnrealNAS, we also conduct ablation studies with anothter different NAS method SGAS \cite{li2019sgas}. We replace the search dataset with the proposed Unreal datasets and keep all the search and training setting the same as SGAS. The obtained architectures are denoted as UnrealNAS-SGAS-RLRD, UnrealNAS-SGAS-RLGD and UnrealNAS-SGAS-RLRN respectively for architectures derived from different datasets.
In our experiments, we find that searching on Unreal datasets yields comparable results as SGAS (slightly less accuracy with smaller parameter size), which validates the generalization ability of UnrealNAS. We also think that Unreal datasets can be applied on other NAS methods based on reinforcement learning or evolutionary strategies, which we leave as our future work.

\begin{table*}[h]
\centering
\resizebox{0.96\linewidth}{!}{%
\begin{tabular}{lccccc}
\toprule
\textbf{{Architecture}}  & \textbf{Test Err. (\%)} & \textbf{Params}  &\textbf{Search}&\textbf{{Search}}  \\
&                            \textbf{CIFAR10} & \textbf{(M)} & \textbf{Data} &\textbf{Label} \\
\midrule
DARTS \cite{Zela2020Understanding} & 2.91$\pm$0.25 & 3.3 & CIFAR10 & Real Label\\ 
DARTS \cite{liu2018darts} & 3.0$\pm$0.14 & 3.3 & CIFAR10 & Real Label\\
UnrealNAS-DARTS-RLRD & 2.62(2.71$\pm$0.08) & 3.0 & CIFAR10 & Random Label \\
UnrealNAS-DARTS-RLGD & 2.46(2.62$\pm$0.10) & 4.3 & FracDB-100(generative) & Random Label  \\
UnrealNAS-DARTS-RLRN & 2.61(2.73$\pm$0.12) & 3.0 & Gaussian Noise & Random Label  \\
\midrule
SGAS \cite{li2019sgas} & 2.39(2.66$\pm$0.24) & 3.7 & CIFAR10 & Real Label \\
UnrealNAS-SGAS-RLRD & 2.74(2.9$\pm$0.09) & 3.1 & CIFAR10 & Random Label \\
UnrealNAS-SGAS-RLGD & 2.44(2.6$\pm$0.11) & 4.0 & FracDB-100(generative) & Random Label  \\
UnrealNAS-SGAS-RLRN & 2.71(2.84$\pm$0.09) & 2.9 & Gaussian Noise & Random Label  \\
\bottomrule
\end{tabular}}
\vspace{5pt}
\caption{\textbf{Comparison with NAS results on CIFAR10.} 
The average and best performance of UnrealNAS are reported.}
\savespace
\label{tbl:compare_different_search_methods}
\end{table*}

\section{Broad Impact}\label{supp:additionalexperiment}
Deep learning has brought great success to various tasks such as computer vision and natural language processing, improving the life of human beings. However, existing deep neural networks require a lot of human effort to design the network architecture and annotate the training data. In this paper, we design an automatic way to search the optimal network architecture with three types of unreal data: 1) only randomly labeled images; 2) generated images and labels; and 3) generated Gaussian noise with random labels. Our method achieves competitive results without using real data or labels, which significantly saves the cost and human effort to collect and annotate data. Besides, our method can also help protect the privacy of the training data used for NAS. These advantages enable our method to have significant practical impact, because the data or annotations may not be available in many applications due to privacy or cost constraint. The potential negative societal impact of our work can be the energy cost for searching the network architectures. However, our method has already shown more efficient searching process than the baselines. Additionally, there are existing methods~\cite{guo2020single,dong2020hawq,cai2020zeroq} that can make training and inference of networks more efficient, which can be integrated into our searching pipeline. We will keep improving the efficiency in the future work.